\documentclass[final,12pt]{clear2026} % Final submission
\usepackage{times}
\usepackage{pslatex}
\usepackage{hyperref}
\usepackage{url}
\usepackage{graphicx}
\usepackage{xcolor}
\usepackage{amsmath}
\usepackage{amsfonts}
\usepackage{lineno}
\usepackage{subcaption}
\usepackage{multirow}
\usepackage{longtable}
\usepackage{makecell}
\usepackage{etoolbox}
\usepackage{pdflscape} % landscape view for table
\usepackage{array} % vertical centering for multirow cells

\title[Bayesian Ablation]{Understanding Task Representations in Neural Networks via Bayesian Ablation}
 
\clearauthor{%
 \Name{Andrew J. Nam} \Email{andrewnam@princeton.edu}\\
 \addr Princeton Laboratory for Artificial Intelligence, Princeton University
 \AND
 \Name{Declan Campbell} \Email{idcampbell@princeton.edu}\\
 \addr Princeton Neuroscience Institute, Princeton University
 \AND
 \Name{Thomas L. Griffiths} \Email{tomg@princeton.edu}\\
 \addr Department of Psychology, Princeton University
 \AND
 \Name{Jonathan D. Cohen}\thanks{Equal contribution; authors listed alphabetically} \Email{jdc@princeton.edu}\\
 \addr Princeton Neuroscience Institute, Princeton University
 \AND
 \Name{Sarah-Jane Leslie}\footnotemark[1] \Email{sjleslie@princeton.edu}\\
 \addr Department of Philosophy and Center for Statistics and Machine Learning, Princeton University
}

\begin{document}
% \linenumbers

\maketitle

\begin{abstract}%
    Neural networks are powerful tools for cognitive modeling due to their flexibility and emergent properties. 
    However, interpreting their learned representations remains challenging due to their sub-symbolic semantics. 
    We introduce a novel probabilistic framework for interpreting latent task representations in neural networks. 
    Inspired by Bayesian inference, our approach defines a distribution over representational units to infer their causal contributions to task performance. 
    Using ideas from information theory, we propose a suite of tools and metrics to illuminate key model properties, including representational distributedness, manifold complexity, and polysemanticity.
\end{abstract}


\section{Introduction}

\renewcommand{\thefootnote}{}\footnotetext{Code available at \url{https://github.com/andrewnam/bayesian_ablation_clear}}\renewcommand{\thefootnote}{\arabic{footnote}}

Neural networks have long been used as tools for understanding human cognition \citep{rumelhart1986general}, from minimalist architectures with just 12 learnable weights \citep{cohen1990control} to large-scale language models such as GPT-5 \citep{openai_gpt5_system_card_2025} with over hundreds of billions of parameters that exhibit human-like cognitive biases and irregularities \citep{binz2023using, binz2024centaur, lampinen2024language, webb2023emergent}.
As these models grow increasingly complex, however, their underlying representations and processes become more opaque, with mechanistic interpretation restricted to simpler architectures such as linear networks \citep{saxe2019mathematical} and attention-only transformers \citep{olsson2022context}. 
This challenge is pronounced in interpreting latent representations of tasks, especially as language models approach limitless capacity for learning tasks and domains described in natural language \citep{bubeck2023sparks, yu2023skill}.

However, a key limitation of inspecting representational layers is that they provide only observational evidence, so causal claims require manipulation \citep{holland1986statistics}.
Inspecting raw network activations provides only a snapshot of representations, mixing task-relevant and task-irrelevant variability \citep{belinkov2022probing} without revealing which elements are causally necessary for task success \citep{nam2025causalheadgating}.
Thus, popular measures that rely on representational vectors such as Euclidean or cosine distances fail to distinguish between values that reflect task-relevant similarity and merely incidental values.
Even direct manipulation of a set of representational units offers limited causal insight without assessing counterfactuals that test whether different representations could lead to similar outcomes \citep{pearl2009causality}.
This becomes especially pronounced in the presence of higher-order dependencies when a behavioral disruption from one ablation can be reversed by applying a second concurrent ablation \citep{fakhar2022systematic}.

We address this issue by applying a Bayesian perspective which identifies causal task representations given an ablation pattern over elements in a neural network, which we represent using a binary mask vector that indicates which elements to lesion by setting their values to 0.
Unlike traditional ablation studies that evaluate \( P(\text{correct} \mid \text{task, mask}) \) to assess how task performance changes when specific representational units of a model are ablated, our approach computes its Bayesian posterior.
The resulting distribution \( P(\text{mask} \mid \text{task}, \text{correct}) \), which we call the ablation mask distribution (AMD), defines a distribution over masks conditioned on correct task performance to describe which units are most likely to have contributed to successful model outputs.
This posterior formulation enables causal reasoning about task representations by explicitly modeling the statistical dependencies between representational units and task outcomes: if a specific set of units is crucial for a task, the probability of masking them given task success will be low.

While prior works have similarly explored exhaustively ablating all possible combinations of network activations \citep{fakhar2022systematic}, our approach neatly integrates causal representational analysis into a single probabilistic object that is mathematically motivated and amenable to further theoretic analyses.
Beyond interpreting individual unit roles, AMD enables summarizing and quantifying broader representational structure using information theory.
Moreover, because the AMD captures higher-order interactions and complex manifold geometry, it supports interpretability without imposing architectural assumptions or constraints.

Our work makes three primary contributions. 
First, we introduce ablation mask distributions as a rigorous and theoretically motivated framework for uncovering causal mechanisms in neural networks that enable the application of information-theoretic tools to study task representations. 
Second, we demonstrate the application of AMDs in a simple, well-understood model, showing how it fits into and extends existing approaches for analyzing representational structure.
We validate our analyses where possible by comparing them to widely-used metrics such as the \(L_1\)-norm or cosine similarity to verify that our measures capture similar regularities as observational measures while guaranteeing that the observed relations are truly causal.
Third, we present a computationally efficient approximation method that addresses the limitations of exhaustively computing the full Bayesian posterior, enabling scalable application of our framework to larger models.

We begin by defining the distribution over ablation masks and its relationship to the model's task performance. 
Next, we demonstrate our approach by applying it to the Integrated Semantics and Control (ISC) model \citep{giallanza2024integrated}, a simple feed-forward multitask neural network trained on human-rated semantic data designed to investigate emergent semantic cognition in context-switching scenarios.
We selected the ISC model for its alignment with human responses on measures such as context similarity and for its architectural simplicity, which facilitates the application and validation of novel methods.
Using this model, we first analyze the exact AMD to characterize several representational properties, including distributedness, manifold complexity, task representational similarity, and task polysemanticity through reverse inference.
We then introduce an approximation method that reduces the computational cost of estimating the full AMD.
Finally, we discuss the limitations of our method and outline potential directions for future work.

\section{Methods}

\subsection{Integrated Semantics and Control (ISC) model} 

\begin{figure}[h]
  \centering
  \begin{minipage}[t]{0.5\linewidth}
    \centering
    \includegraphics[width=\linewidth]{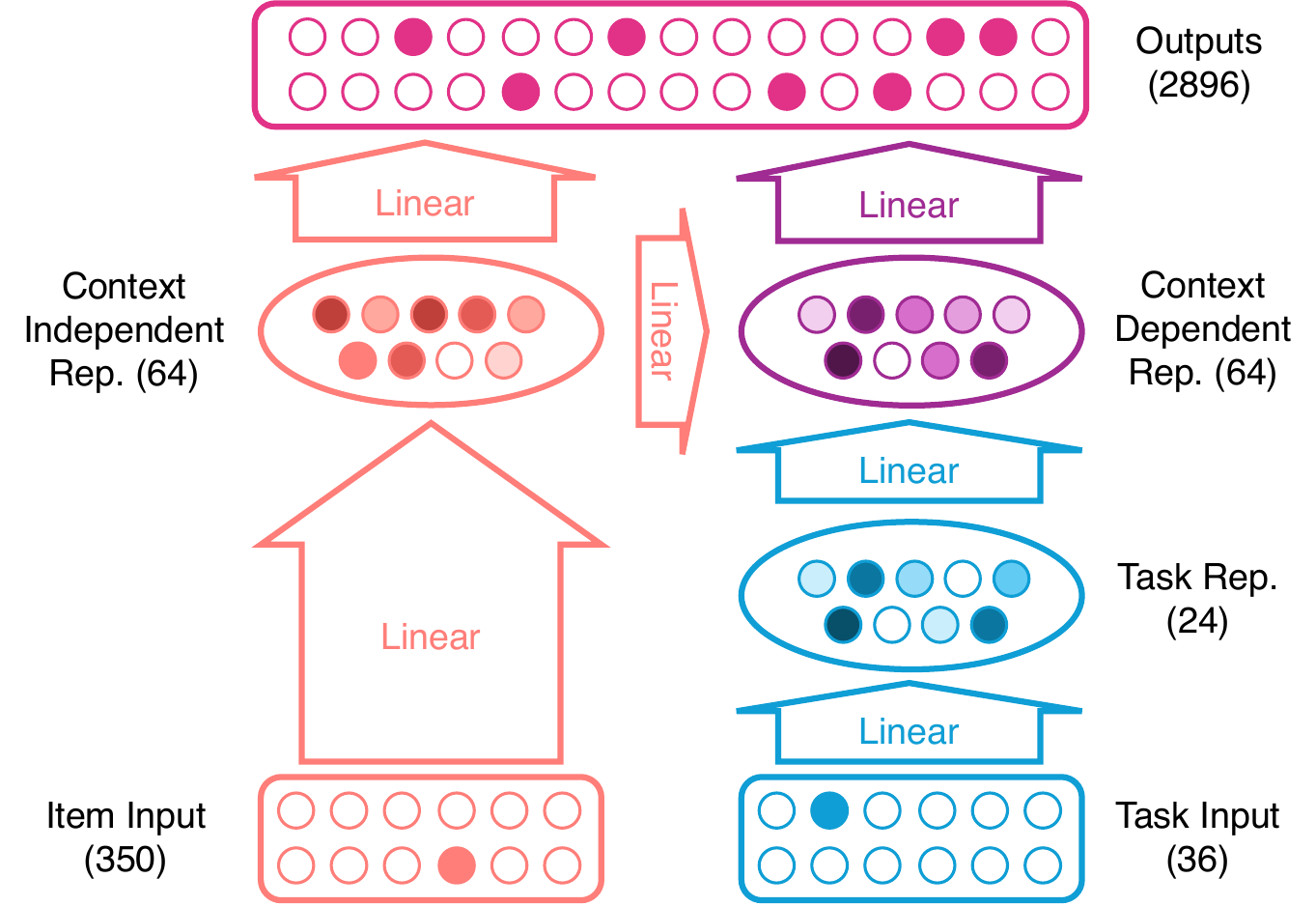}
    \caption*{(a)}
  \end{minipage}
  \hfill
  \begin{minipage}[t]{0.44\linewidth}
    \centering
    \includegraphics[width=\linewidth]{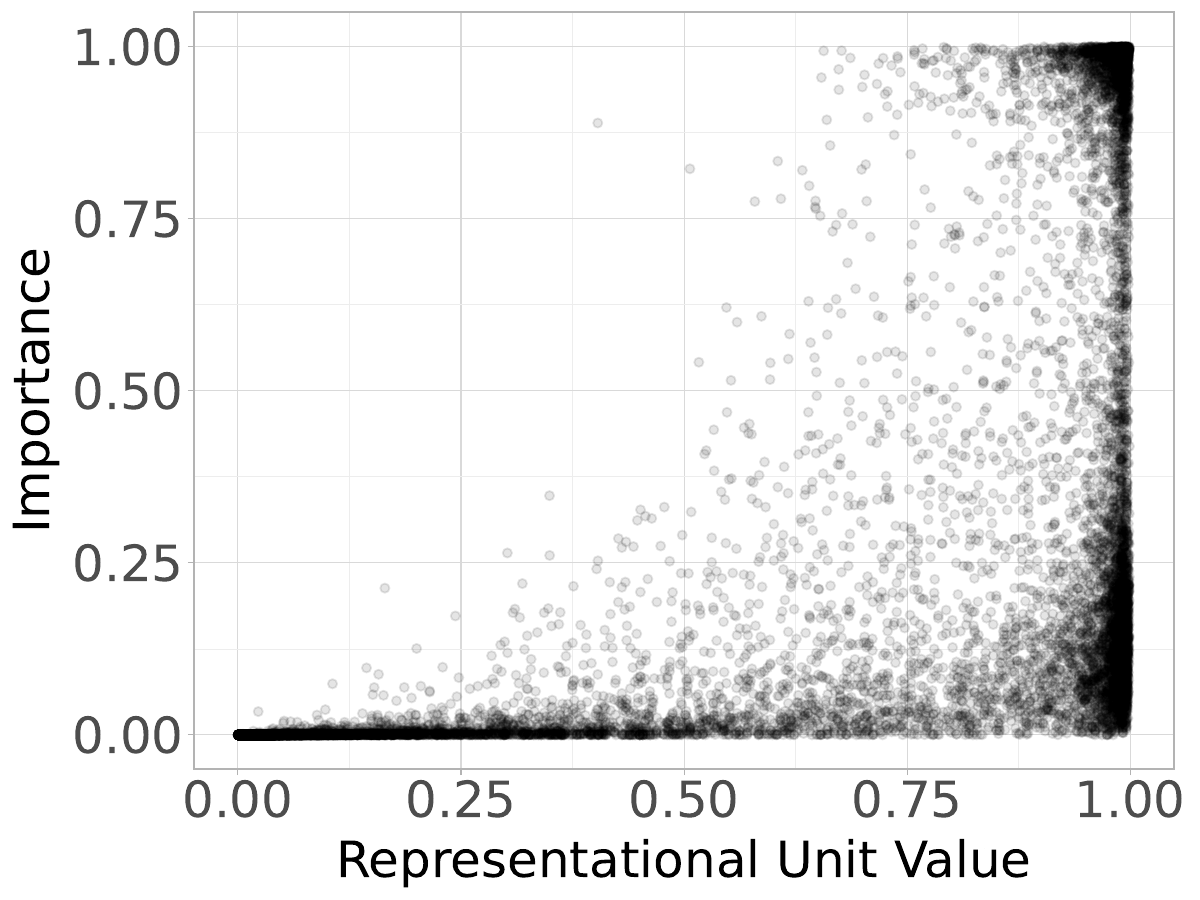}
    \caption*{(b)}
  \end{minipage}
\caption{(a) ISC model. Number of representational units shown in parentheses. 
% Sigmoid activation function is applied after each linear layer. 
(b) Task representation unit values \(h_i(t)\) and their importance \(1 - H(m_i \mid t,c)\).}
\label{fig:isc_importance}
\end{figure}

The Integrated Semantics and Control (ISC) model (Figure~\ref{fig:isc_importance}a) is trained on the Leuven Concepts Database \citep{de2008word, ruts2004dutch, storms2001flemish}, a human-rated semantics dataset containing 350 animals and 2,896 features that are grouped into 36 distinct feature classes \citep{wu2009perceptual} (see SI Table \ref{table:dataset_property_types}). 
The model is trained to simultaneously predict the features of a particular animal (item input) and a subset of its features within a feature class (task input).
For instance, giving the model the animal ``elephant'' and the ``category'' feature class would produce positive outputs only for features relevant to an elephant's category, e.g. ``is an animal''.

We adopt the architecture described by \citet{giallanza2024integrated}. The inputs, item (animal) and task (feature class), are represented as one-hot vectors (i.e. a vector of 0s with a single 1), which are mapped to separate embedding spaces: the context-independent representation layer and the task representation layer.
The context-independent layer is used to directly predict all features of the animal. It also provides input to the context-dependent layer, along with the task representation, which together form the context-dependent representation.
Notably, the task representation, which we apply our ablations to, modulates the context-independent representations by effectively directing the network's attention to the features of the input that are most relevant to the specified task.
This context-dependent representation is then used to predict the set of features specified by the task input.
We also introduce a null-task during training, represented by a zero-vector embedding and a zero-vector target output, which effectively encourages the model to learn strongly negative output biases and more structured embeddings across the 36 feature classes.

In this paper, we apply our method to the task representation layer of the ISC model. 
Although using a model with an explicit task representation might seem to limit the validity and generality of an approach designed to be applicable even to models without such representations, this choice serves a critical purpose. Starting with a model that has clearly defined task representations allows us to rigorously evaluate a novel approach in a controlled environment, where the model’s properties are well understood. By first validating the approach in this setting, we aim to establish a firm foundation for extending these tools to provide new insights in less interpretable systems.

\subsection{Ablation mask distribution}  
We define the ablation mask distribution 
\( P(\text{mask} \mid \text{task}, \text{correct}) \) 
as a conditional distribution over binary vectors that mask representational units of a neural network. If the mask value is 0, the representational unit is replaced with zero; if the value is 1, the unit is left unchanged. 

In our experiments, we apply these masks to the task representation layer of the ISC model, multiplying the binary mask vector \( m \in \{0, 1\}^d \) element-wise by the post-activation values of the task representation units \( h \in (0, 1)^d \), where \( h \) follows a sigmoid activation function.
We measure model performance conditioned on a task and an ablation mask, \( P(\text{correct} \mid \text{task}, \text{mask}) \), by applying the mask and taking the feature predictions for all 350 animals. A prediction is mapped to \texttt{true} if the predicted feature likelihood in the output layer is \( \geq 0.5 \) and \texttt{false} otherwise.  
Each prediction is compared against the target value in the data to determine whether or not it is correct.

Given the highly skewed distribution of positive and negative feature values in the dataset, we estimate task-mask performance using the geometric mean of the model’s sensitivity and specificity:  
\[
P(\text{correct} \mid \text{task}, \text{mask}) =
    \sqrt{\smash[b]{%
    \underbrace{P(\text{correct} \mid \text{task}, \text{mask}, \text{target} = 1)}_{\text{sensitivity}}}}
    \times
    \sqrt{\smash[b]{%
    \underbrace{P(\text{correct} \mid \text{task}, \text{mask}, \text{target} = 0)}_{\text{specificity}}}}
\]
% \begin{equation*}
% \begin{split}
%     P(\text{correct} \mid \text{task}, \text{mask}) =
%     \sqrt{\smash[b]{%
%     \underbrace{P(\text{correct} \mid \text{task}, \text{mask}, \text{target} = 1)}_{\text{sensitivity}}}}
%     \times
%     \sqrt{\smash[b]{%
%     \underbrace{P(\text{correct} \mid \text{task}, \text{mask}, \text{target} = 0)}_{\text{specificity}}}}
% \end{split}
% \end{equation*}
% \\
\\
\noindent The inclusion of the null-task described above, which encourages the model to predict 0 for all features, and the geometric mean, which ensures balanced evaluation of positive and negative features, results in a 0\% ``chance'' accuracy on all feature classes. 

For brevity, we denote the ablation mask as \( m \), the task as \( t \), and the correctness indicator as \( c \). 
Using this notation, the correctness probability serves as the basis for defining the ablation mask distribution with Bayes' rule:
\[
P(m \mid t, c) = \frac{P(t \mid m, c) P(m \mid c)}{\sum_{m'} P(t \mid m', c) P(m' \mid c)}
\]
This distribution over ablation masks identifies the subset of causally relevant units that allow the model to successfully perform a specific task, offering a principled framework for interpreting the functional contributions of representational units within neural networks.

While the Bayesian formulation of the correctness metric is mathematically motivated, its separation between high and low performance can lead to a posterior distribution that is too flat to be useful in practice. 
For instance, in a task with a baseline accuracy of 50\%, a ``success'' mask achieving 95\% accuracy would be sampled only about twice as often as a ``failure'' mask that performs considerably worse (i.e. reduces performance to chance).
This distribution can result in an unbalanced exploration of mask space, potentially making it harder to interpret the functional contributions of different units.

Instead, we convert accuracy measures $p$ into odds-ratios \( \frac{p}{1-p} \), which amplify performance differences, so that a mask with accuracy $p=0.95$ is sampled approximately 20 times more often than one with accuracy $p=0.50$.
This aligns naturally with the sigmoid nonlinearity inherent in the model, given by 
\[
\sigma(x) = \frac{1}{1 + e^{-x}} \quad \quad \quad x = \log\left(\frac{p}{1-p}\right).
\]
The sigmoid function serves as the inverse of the log-odds transformation, mapping log-odds into probabilities. 
% Focusing on the odds-ratio thus measures the impact of the mask on the input to the sigmoid.
Assuming a uniform prior over the ablation masks, the distribution of the mask given the task and correctness is expressed as
\[
P(m \mid t, c) = 
\underbrace{\vphantom{\left( \sum_{m'} \frac{P(c \mid t, m')}{1 - P(c \mid t, m')} \right)^{-1}} \frac{P(c \mid t, m)}{1 - P(c \mid t, m)}}_{\text{Likelihood}}
\cdot
\underbrace{\left( \sum_{m'} \frac{P(c \mid t, m')}{1 - P(c \mid t, m')} \right)^{-1}}_{\text{Normalization}}
\]

\noindent We find that the odds-ratio modification aligns with other measures describing model representation and behavior better than using the standard Bayesian formulation (see SI).

\section{Analyses} 
While the central advantage of the AMD is its ability to distinguish between causally relevant and merely incidental activations of representational units, it also has a secondary advantage of reformulating task representation vectors into probability distributions, which opens a suite of various analytic tools.
In this section, we explore the application of information-theoretic measures and optimal transport theory to quantify several properties of the task representations.
All analyses were performed on 10 separate instances of the ISC-model.

\subsection{Entropy}
We begin our analyses by considering the entropy $H(m \mid t, c)$ of the ablation mask distribution, which measures the diversity of masks that are sufficient to perform well on a task.
\begin{equation*}
    H(m \mid t, c) = 
    - \sum_{m} P(m \mid t, c) 
    \cdot \log P(m \mid t, c)
\end{equation*}

To illustrate this relationship, consider three types of task representation units: (1) units necessary for task performance, (2) units that interfere with task performance, and (3) units that are irrelevant to the task. 
Suppose, for instance, that a unit \( h_1 \) must remain near 1 for the model to perform well, and ablating it (setting it to 0) reduces task performance to chance; in this case, the marginal probability \( P(m_1 \mid t, c) \) would be near 1. 
Conversely, if a unit \( h_2 \) interferes with performance, its activation would reduce accuracy to chance, and \( P(m_2 \mid t, c) \) would be near 0. 
Both \( h_1 \) and \( h_2 \) are thus causally relevant to the task and favor specific mask values (1 and 0 respectively), increasing the concentration of the ablation mask distribution. 
In contrast, an incidental unit \( h_3 \) that does not significantly affect outcomes will have \( P(m_3 \mid t, c) \) near 0.5, decreasing concentration. 
Since each incidental unit contributes an independent binary choice (0 or 1) without affecting correctness, the number of high-probability masks grows exponentially with the number of incidental units, thereby increasing entropy. 
Thus, higher entropy reflects task representations with fewer causally relevant units and more incidental units.

Similarly, the entropy of an individual representational unit $h_i$ over a specific task reflects how strongly the model depends on the particular unit's value. 
To compute this, we start with the marginal probability $P(m_i \mid t, c)$, which represents the likelihood of the unit $h_i$ being active (not ablated) during successful task performance.
We then measure the the marginal entropy \(H(m_i \mid t, c)\), which is bounded between 0 and 1, using \(p = P(m_i \mid t, c)\).
\[
P(m_i \mid t, c) = \sum_m m_i \cdot P(m \mid t, c)
,\quad 
H(m_i \mid t, c) = -p \log p - (1 - p)\log(1 - p).
\]
% \[
%     P(m_i \mid t, c) = \sum_m m_i \cdot P(m \mid t, c).
% \]
% Letting \(p = P(m_i \mid t, c)\), the marginal entropy \(H(m_i \mid t, c)\), bounded between 0 and 1, is
% defined as
% \[
%     H = -p \log p - (1 - p)\log(1 - p).
% \]

\subsubsection{Unit importance and representational distributedness \label{subsubsec:unit_importance}}
Representational units contribute to the formation of distributed representations \citep{hinton1986distributed}, a topic extensively studied in both large \citep{arora2018linear, bricken2023monosemanticity} and small \citep{hinton1986learning} models.
Although the representational unit value and a distributedness metric such as the \(L_1\)-norm of a task representation vector are easy to access and compute, these observational measures permit at most correlational interpretations and cannot provide causal interpretations.
Here, we leverage the AMD to compute an analog of a representational unit value and representational distributedness with a causal interpretation, i.e. which units are actually relevant to the task.

We measure the model's reliance on the representational unit $h_i$ using \( 1 - H(m_i \mid t, c) \) (where H is bounded between 0 and 1), which we refer to as unit importance.
While this measure is correlated (\( r = 0.661\)) with the actual representational unit's value \( h_i(t) \), its deviation indicates when the representational unit is not strictly necessary for the model to perform well.
As shown in Figure~\ref{fig:isc_importance}b, while unit importance generally increases with \( h_i(t) \), a large number of representational units reflect low importance despite high activation values.

\begin{figure}[h]
  \centering
  \begin{minipage}{0.49\linewidth}
    \centering
    \includegraphics[width=\linewidth]{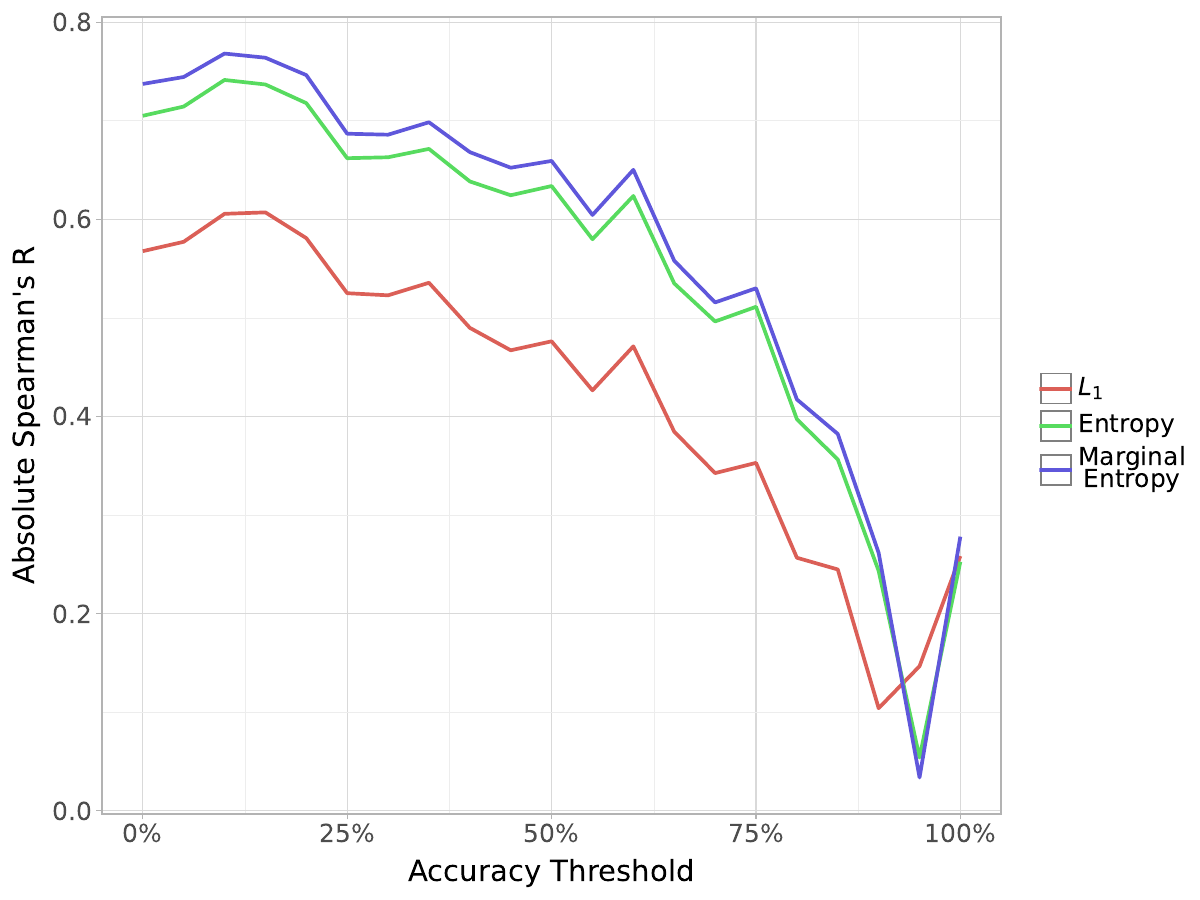}
    \caption*{(a)}
  \end{minipage}
  \hfill
  \begin{minipage}{0.49\linewidth}
    \centering
    \includegraphics[width=\linewidth]{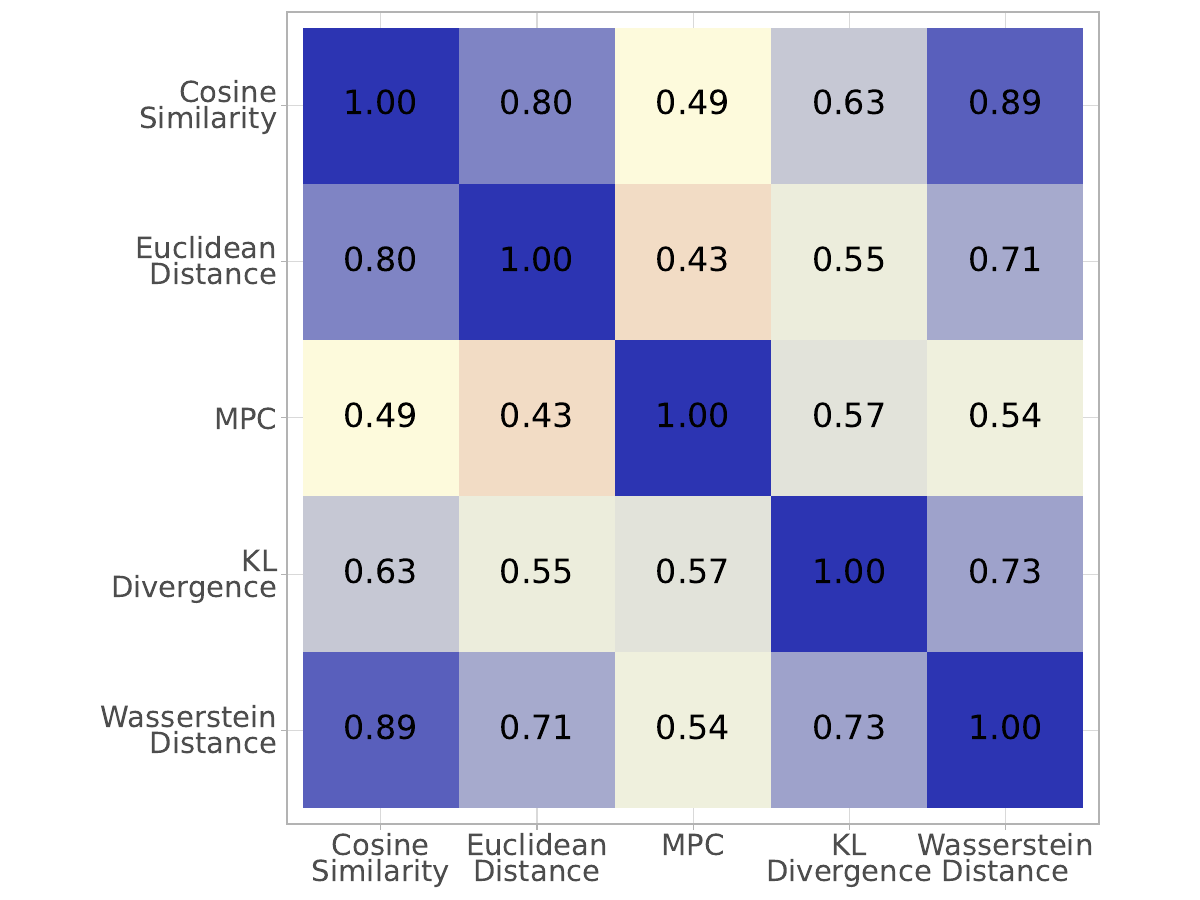}
    \caption*{(b)}
  \end{minipage}
  \caption{(a) Correlation between task representation metrics and task acquisition order across accuracy thresholds. (b) Spearman correlation between similarity measures.}
  \label{fig:task_rank_rsa}
\end{figure}

% \begin{figure}[h]
%   \centering
%   \includegraphics[width=0.4\textwidth]{figures/importance_h24.pdf}
%   \caption{Task representation unit values $h_i(t)$ and their importance \(1 - H(m_i \mid t,c) \).}
%   \label{fig:importance}
% \end{figure}

% \begin{figure}
%   \centering
%   \includegraphics[width=0.4\textwidth]{figures/task_rank_correlation.pdf}
%   \caption{Correlation between and task representation metrics and task acquisition order along different accuracy thresholds. Task acquisition order is defined as the order that a model's task accuracy first exceeds the specified threshold.}
%   \label{fig:task_rank}
% \end{figure}

As might be expected, tasks with more distributed representations rely on a greater number of representational units with high importance, whereas more localized representations concentrate importance on only a few units. 
Thus, marginal entropy also provides a way to measure the effective representational distributedness of a task across the \( d \) representational units. 
We quantify this by summing across all \( d \) representational units: \( d - \sum H(m_i \mid t, c) \).
Naturally, this relates to the \(L_1\)-norm of the activation values (\(\sum |h_i(t)|\), \(r = 0.922\)), but with the guarantee of a causal, rather than merely observational or correlational, interpretation.
 
\subsubsection{Manifold complexity \label{subsubsec:manifold_complexity}}
The distributed representations of neural networks often reflect dependencies, such that the activation of some units change how other units are used by the network by warping the representational manifolds \citep{fakhar2022systematic, giallanza2024integrated}.
We quantify this manifold complexity by measuring the joint entropy \( H(m \mid t, c) \) of the AMD, which captures the full statistical dependencies between representational units, offering a more holistic view than the marginal entropy sum \( \sum H(m_i \mid t, c) \). 
Comparing these two entropic measures allows us to quantify the information contained in higher-order dependencies, expressed as a normalized entropy drop:
\[
\Delta H = 1 - \frac{H(m \mid t, c)}{\sum_i H(m_i \mid t, c)}.
\]
The value of \( \Delta H \) represents the proportion of entropy attributed to higher-order dependencies, providing a measure of the manifold complexity of task representations.

In the ISC model, we observe an average entropic reduction of $\Delta H = 4.62\%$, indicating that task representations are predominantly modular. 
This suggests minimal reliance on higher-order interdependencies among units, which may be expected for a simple feed-forward network designed for a specific set of semantic cognition tasks.

\subsubsection{Task differentiation}

% \begin{figure}
%   \centering
%   \includegraphics[width=0.4\textwidth]{figures/task_rank_correlation.pdf}
%   \caption{Correlation between and task representation metrics and task acquisition order along different accuracy thresholds. Task acquisition order is defined as the order that a model's task accuracy first exceeds the specified threshold.}
%   \label{fig:task_rank}
% \end{figure}

Progressive differentiation of internal representations is a fundamental property of neural networks that offers a unifying framework for developmental psychology and statistical learning \citep{mcclelland2003parallel, munakata2003connectionist}.
However, existing tools for describing these phenomena through mathematical theory are often limited to certain architectural constraints, such as only containing linear layers \citep{saxe2019mathematical, lampinen2018analytic}.
Because AMD captures full statistical dependencies, we hypothesized that it would be well-suited for analyzing training dynamics in nonlinear systems.
Applied to the ISC model, we 
% hypothesized that causal relevance and distributedness of the task representation units would be related to how the model differentiates its representations during training.
% Specifically, we 
posited that as the task-representations differentiate from the null-task, 
tasks learned earlier during training would recruit more representational units.
Conversely, tasks learned later would require fewer units as they incrementally diverge from established representations.
% represented by a zero-vector embedding and target output, would lead 
% to differentiate themselves from the zero-vector by raising 
% Conversely, tasks learned later would require fewer units as they incrementally diverge from established representations.

To test this hypothesis, we recorded the order in which the accuracy first rose above 0\% for each task, which corresponds to the initial accuracy for all tasks due to the presence of the null-task.
We then compared this rank metric to the two entropic measures and the \(L_1\)-norm of each task representation using Spearman's rank-order correlation.
Consistent with our hypothesis, we found a high absolute correlation between the order that the model learned each task and the two entropy measures ($r = 0.708$ for joint entropy, $r = 0.746$ for marginal entropy).
Despite its relatively high correlation with entropy, the \(L_1\)-norm had only a moderate correlation of $r = 0.573$ with the task acquisition order.
This indicates that the ablation masks' capacity to distinguish between causally relevant and merely incidental representational values may allow them to capture model properties more accurately. 
Moreover, when we compared the correlation using various accuracy thresholds as shown in Figure~\ref{fig:task_rank_rsa}a, we found that the absolute correlation begins to drop around an accuracy threshold of 15\%, suggesting that the entropic measures are sensitive to how the model initially allocates representational units but less to how it refines their values during training.

\subsection{Mutual information, reverse inference, and polysemanticity}
We now consider how the AMD can be used to address the reverse inference problem of identifying the task from the representational unit activations.
That is, beyond evaluating whether a representational unit $h_i$ contains sufficient information to decode the task, we ask whether $h_i$ is \emph{causally involved} in encoding the task, i.e. whether the unit is mono- or polysemantic \citep{arora2018linear, elhage2022superposition, bricken2023monosemanticity} with respect to the various tasks in the dataset.

By leveraging the AMD, we isolate $h_i$'s necessity for task performance, distinguishing incidental activations from task-relevant contributions and quantifying how its influence is distributed across multiple tasks.
This conditional probability $P(t \mid m, c)$ and its marginalization over individual units $P(t \mid m_i, c)$, where $m'_i$ is the value of the $i^\text{th}$ bit in mask $m'$, are given by
\[
P(t \mid m, c) = \dfrac{P(m \mid t, c) \cdot P(t \mid c)}{\sum_{t'} P(m \mid t', c) \cdot P(t' \mid c)}
,\quad 
P(t \mid m_i, c) = \frac{\sum_{m'} m'_i \cdot P(t \mid m', c) \cdot P(m', c)}{\sum_{t', m'} m'_i \cdot P(t' \mid m', c) \cdot P(m', c)}
\]

Using the conditional task distribution, we compute the mutual information by measuring the reduction in entropy, which we normalize for interpretability
\[
I_n(t, m \mid c) = 1 - \dfrac{H(t \mid m, c)}{H(t \mid c)} 
,\quad 
I_n(t, m_i \mid c) = 1 - \dfrac{H(t \mid m_i, c)}{H(t \mid c)}
\]

The entropy of the task distribution conditioned on unit activation provides a measure of task polysemanticity.
For example, consider a representational unit that encodes exactly one task $t'$, so that $P(t' \mid m_i, c) = 1$ and 0 for all other tasks.
In this case, the resulting entropy $H(t' \mid m_i, c)$ and the normalized mutual information $I_n(t', m_i \mid c)$ are 0 and 1 respectively.
Conversely, a unit that provides no task information when considered independently, so that $P(t \mid m_i, c) = P(t)$, will result in $I_n(t, m_i \mid c) = 0$.
Thus, $I_n$ provides a bounded measure of a unit's task specificity, ranging from 0 to 1.
While \(I_n = 1\) indicates perfect determinism and \(I_n = 0\) indicates no task relevance, any \(I_n > 0\) indicates that a unit encodes some useful information, just not in isolation. 
Instead, its contribution must be combined with others, either conjunctively or compositionally, to represent the full task structure.
Thus, this measure reflects polysemanticity with respect to tasks rather than sub-task concepts, as a unit with low \(I_n\) for tasks may still encode meaningful sub-task structure, e.g. semantic features across multiple tasks, allowing it to participate in structured, compositional representations that support task generalization.

We compare the difference between the two measures computed on the full mask distributions and on the marginal unit distributions, and find that individual units share very little mutual information with tasks when considered independently, reducing entropy by about 4.21\% in most cases. 
In contrast, ablation masks are significantly more informative, reducing uncertainty by an average of 82.6\%.
This suggests that the representational units individually encode little information about particular tasks, and that the model relies on ensembles of representational units through distributed representations. 
At the same time, because individual units participate in multiple task representations, this also implies a form of polysemanticity, where units contribute to multiple tasks in a structured way rather than encoding distinct tasks independently.

% \begin{figure}[h]
%   \centering
%   \begin{minipage}{0.48\linewidth}
%     \centering
%     \includegraphics[width=\linewidth]{figures/task_info_gain_h24.pdf}
%     \caption*{(a)}
%   \end{minipage}
%   \hfill
%   \begin{minipage}{0.48\linewidth}
%     \centering
%     \includegraphics[width=\linewidth]{figures/rsa_h24_odds_ratio.pdf}
%     \caption*{(b)}
%   \end{minipage}
%   \caption{(a) Percentage of normalized mutual information captured by the full AMD and by marginal unit distributions; (b) Spearman correlation between similarity measures.}
% \end{figure}

\subsection{Task similarity}
Thus far, we have focused on measures targeted at understanding individual task representations. In this section, we extend our analysis to compare the similarity between task representations. Because the ablation mask distributions reflect causal relevance and higher-order dependencies between representational units, they capture relationships that vector-based measures, such as cosine or Euclidean distance, may overlook. For example, under cosine or Euclidean distance, the vector [1,1] would be considered further from [0,0] than [0,1], even if the second dimension is not meaningfully used by the model.
To compare the task similarities using the ablation mask distributions, we turn to two distance measures well-suited for comparing probability distributions: KL-divergence and Wasserstein distance.

The KL-divergence $D_{KL}(P \| Q)$ is a widely-used measure in information theory for quantifying the difference between two probability distributions by capturing the amount of information lost when approximating one distribution ($Q$) with another ($P$).
However, because $D_{KL}(P \| Q)$ is inherently asymmetric, we use the symmetrized KL-divergence \(D^S_{KL}(P \| Q)\) which combines $D_{KL}(P \| Q)$ and $D_{KL}(Q \| P)$ into a bidirectional measure.
\[
D_{\text{KL}}(P \| Q) = \sum_{x} P(x) \log \frac{P(x)}{Q(x)}
,\quad 
D^S_{\text{KL}}(P \| Q) = \frac{1}{2} D_{\text{KL}}(P \| Q) + \frac{1}{2} D_{\text{KL}}(Q \| P)
\]

We also consider the Wasserstein distance $W(P,Q)$, which quantifies the minimal cost of transforming one distribution into another
\begin{equation*}
    W(P, Q) = \inf_{\gamma \in \Gamma(P, Q)} \mathbb{E}_{(x, y) \sim \gamma} [d(x, y)],
\end{equation*}
where \( \Gamma(P, Q) \) is the set of joint distributions (couplings) with marginals \( P \) and \( Q \), and \( d(x, y) \) is the Hamming distance between mask configurations \( x \) and \( y \), i.e., the number of 1's or 0's that need to be flipped to transform one mask into another.

A key difference between KL-divergence and Wasserstein distance is that the latter is informed by the distance metric (Hamming distance) while KL-divergence is not. 
For example, if two distributions agree on all but two masks, the KL-divergence between them will depend only on the differing amounts of mass placed on these masks. Wasserstein distance is sensitive to this difference in mass, and also to the Hamming distance between the masks. In particular, the Wasserstein distance will be greater if the two masks share fewer bits in common, whereas the KL-divergence is not sensitive to this.

To contextualize these probabilistic measures, we compare them to more conventional vector-based metrics: cosine similarity and Euclidean distance. 
While cosine similarity and Euclidean distance do not account for higher-order dependencies or causal relevance, they are reasonable points of comparison as they are used widely in assessing the similarity of vector-based representations, both in neural networks and empirical neural data \citep{kriegeskorte2008representational}.  Furthermore, they are useful  in the evaluating representations in the ISC model, given the relative simplicity of its representational manifold as evidenced by the low entropy drop $\Delta H$.
Additionally, we introduce a non-parametric measure of task similarity that we refer to as mask-performance correlation (MPC), which compares the correlation between the accuracies of two tasks when the same mask is applied. 
This measure provides a direct link between ablation masks and task performance without incorporating the importance weighting involved in probabilistic measures computed over the posterior distribution.
Specifically, MPC measures correlation using $P(c | t, m)$ without further weighting, whereas the AMD metrics weight the distances by $P(m | c, t)$

To compare the various measures, we conduct a representational similarity analysis (RSA) \citep{kriegeskorte2008representational}, computing the absolute Spearman correlation between metrics across task pairs (Figure~\ref{fig:task_rank_rsa}b). KL-divergence and Wasserstein distance exhibit strong correlation ($r = 0.73$), highlighting their shared reliance on posterior mask distributions. 
The especially high correlation ($r = 0.89$) between Wasserstein distance and cosine similarity and suggests that the probabilistic framework preserves much of the structural information captured by traditional similarity metrics. 
Both measures align with intuitive similarities between certain tasks, such as between `lexical expressions' and `synonyms', or between `related actions' and `external features'.
In contrast, MPC exhibits a relatively weak correlation with other measures. For instance, its correlation with cosine similarity drops to $r = 0.50$, suggesting that posterior weighting is important for preserving representational fidelity, beyond merely considering the outcome of performance for each task.

\section{Approximating the Ablation Mask Distribution}

\begin{figure*}[h]
  \begin{center}
  \includegraphics[width=\textwidth]{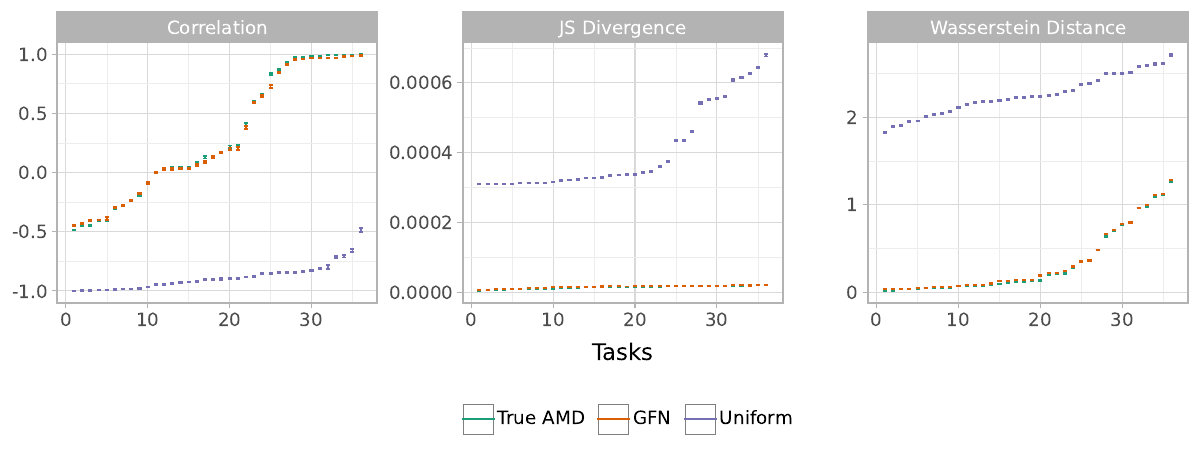}
  \end{center}
  \caption{Comparison between the true AMD, the GFlowNet approximation, and the uniform baseline from a single model seed. Tasks are sorted by the average metric value, and 95\% bootstrapped confidence intervals (CIs) are shown as error bars. Note that the true AMD is often visually obscured beneath the GFlowNet curve.}
  \label{fig:gfn_gof}
\end{figure*}

While exact Bayesian inference provides a principled framework for understanding task representations, the computational cost is prohibitive for large-scale models. The number of possible ablation masks grows exponentially with the number of representational units. Even for our small model with a 24-dimensional task representation layer, computing the full posterior over 36 tasks requires nearly 100 GPU-hours. Thus, exact computation becomes infeasible for larger models, necessitating efficient approximation methods.

A popular method for approximating complex distributions in Bayesian models is Markov Chain Monte Carlo (MCMC) \citep{metropolis1953equation}.
However, MCMC methods are ill-suited for approximating the AMD. 
First, because masks are defined over binary vectors and flipping a single unit can cause large, though not necessarily unpredictable, changes in task performance, methods that rely on smooth probability landscapes, such as Metropolis-Hastings and Hamiltonian Monte Carlo \citep{neal2011mcmc}, struggle to explore the space efficiently.
Second, higher-order dependencies between representational units violate the assumptions of sequential update strategies like Gibbs sampling, causing slow mixing \citep{neal1993probabilistic}. Finally, MCMC methods lack the ability to exploit semantic structure and cannot generalize across similar masks \citep{neal2011mcmc}, requiring explicit evaluation of each configuration to accurately approximate the posterior.

We avoid these limitations by approximating the AMD using a Generative Flow Network \linebreak (GFlowNet) \citep{bengio2021flow, bengio2023gflownet}, a framework that combines reinforcement learning (RL) and generative modeling to learn a sampler for structured objects in proportion to a given reward function. 
Like RL, GFlowNets define a stochastic decision process over a state space, where a sequence of discrete actions constructs a final structured object that can be evaluated by a reward function. 
However, instead of maximizing expected reward, GFlowNets learn a policy that samples solutions in proportion to it, ensuring a diverse set of high-reward candidates rather than only the best-performing ones. 
This distinguishes them from standard RL and makes them more similar to generative models, which approximate a distribution, but unlike models trained on a fixed dataset, GFlowNets actively propose and evaluate candidates, making them more akin to MCMC. 
Thus, GFlowNets are particularly well-suited to our setting: while evaluating an ablation mask is straightforward (applying it to the ISC model and measuring task performance) directly generating task-relevant masks is nontrivial.

The GFlowNet model here learns to sample trajectories that construct ablation masks.  
Each trajectory begins with a mask of 1s, denoted as \( m^{(0)} = \mathbf{1} \), and at each step \( j \), the current mask \( m^{(j)} \) is updated by either setting a bit $m^{(j)}_i$ to 0 or terminating the trajectory ($\top$).  
The distribution of terminal masks \( m^{(\text{final})} \) is proportional to the task performance measure, which we define as the reward function:
\[
    R(m, t) = \frac{P(c \mid t, m)}{1 - P(c \mid t, m)}
\]

Our GFlowNet consists of the following components (where \( j \) represents the step in the trajectory and \( t \) denotes the task):
1. A forward policy model, \( P_\theta^f (m^{(j+1)} \mid m^{(j)}; t) \), which selects either a bit to set to 0 or terminates the trajectory;
2. An auxiliary backward policy model, \( P_\theta^b (m^{(j)} \mid m^{(j+1)}; t) \), which learns the reverse policy by adding back ablated units into the representation.
The backward policy is required for training, as the objective balances the forward and backward policy distributions, but is auxiliary because it is not used when sampling.
We train this model with the detailed-balance objective \citep{bengio2023gflownet}, using the termination probability \( P(\top \mid m^{(j)}, t) \) and the reward function to compute the state flow \citep{deleu2022bayesian}. 
The training objective minimizes

\begin{equation*}
    \mathcal{L} = \Bigg( 
    \log \frac{R(m^{(j)}, t)}{R(m^{(j+1)}, t)}
    + \log \frac{P_\theta(\top \mid m^{(j+1)}, t)}{P_\theta(\top \mid m^{(j)}, t)} 
    + \log \frac{P_\theta^f (m^{(j+1)} \mid m^{(j)}; t)}
                 {P_\theta^b (m^{(j)} \mid m^{(j+1)}; t)}
\Bigg)^2
\end{equation*}

To assess how well the GFlowNet approximates the true AMD, we draw 100,000 samples from the GFlowNet and construct an empirical sample frequency distribution. We compare this to the true posterior using three metrics. First, we compute the Pearson correlation, which provides an intuitive measure of alignment but is insensitive to probability scale. Second, we compute the Jensen-Shannon (JS) divergence \citep{lin1991divergence}, a symmetrized version of KL-divergence that remains suitable for distributions with mismatched support—an issue inherent in approximating a \(2^{24}\)-dimensional space from 100,000 samples. However, JS-divergence amplifies discrepancies in low-probability regions, potentially overstating differences. Lastly, we compute the Wasserstein distance, which is robust to both support differences and distortions in low-probability regions. Since computing Wasserstein distance for the full AMD is intractable, we instead estimate it by comparing against a second 100,000-sample frequency distribution, this time drawn from the true AMD. For consistency, we apply the same sampling-based estimation to Pearson correlation and JS-divergence.
% \citep{feydy2019interpolating}

To contextualize the performance of the GFlowNet approximation, we evaluate it against two additional distributions that serve as upper and lower comparison bounds.  
First, we draw a separate set of 100,000 samples from the true AMD, which allows us to measure the expected error due to sampling variability rather than model fit.  
Second, we draw 100,000 samples from a uniform distribution over all masks, which represents an uninformative prior and provides a lower bound for the comparison metrics.  
To account for sampling variability, we bootstrap \citep{efron1979bootstrap} the distribution of each metric by repeating the full evaluation procedure 200 times.

As shown in Figure~\ref{fig:gfn_gof}, all three metrics indicate a strong alignment between the GFlowNet approximation and the true AMD, with differences that are largely imperceptible at a glance. This suggests that the model successfully approximates the AMD with high fidelity.
However, a closer examination of the bootstrap confidence intervals reveals small but statistically significant discrepancies between the GFlowNet and the true AMD.
Across all tasks, the mean estimate from the GFlowNet falls within the 95\% confidence interval of the true AMD in 27.8\% of cases for Pearson correlation, 16.7\% for JS divergence, and 8.33\% for Wasserstein distance.
Additionally, the confidence intervals of the GFlowNet and the true AMD overlap in 44.4\% of cases for Pearson correlation, 41.7\% for JS divergence, and 13.9\% for Wasserstein distance.
Thus, despite requiring only $\sim$1\% of the compute needed for exact inference, the GFlowNet effectively approximates the AMD with high fidelity, though observed discrepancies indicate room for improvement.

\section{Discussion}

In this paper, we have introduced a novel probabilistic framework for studying task representational structure in neural networks. 
Unlike simple ablation which only evaluates downstream effects on task performance, our approach uses a Bayesian perspective that reconstructs task representations as posterior distributions over ablation masks, allowing for causal interpretation of task representations.
This probabilistic approach facilitates the use of tools from information theory and optimal transport, enabling a deeper exploration of task representations that is sensitive to the structure of the representational manifold. For example, measures such as entropy and mutual information can be used to quantify how a neural networks distributes information in complex manifolds.

While we have demonstrated our method using the ISC model, the framework is designed to be broadly applicable across architectures and domains. 
It is flexible with respect to both the evaluation metric and the structure of the model: beyond binary task accuracy, users could define success in terms of prediction confidence in image classification \citep{krizhevsky2012imagenet}, sequence-probabilities in language modeling \citep{achiam2023gpt}, or other task-relevant objectives. 
Similarly, the approach is agnostic to what is being ablated—beyond individual representational units, it can naturally extend to larger functional modules such as convolutional filters \citep{zeiler2014visualizing, yosinski2015understanding} in convolutional networks \citep{krizhevsky2012imagenet} or attention heads \citep{michel2019sixteen, voita2019analyzing, nam2025causalheadgating} in transformer models \citep{vaswani2017attention}. 
We see this flexibility as a key strength of the framework, opening opportunities for future work to explore causal structure at multiple levels of abstraction in modern deep learning systems.

Our framework has several limitations that prompt further research. 
Although the AMD is designed to be broadly applicable across architectures and evaluation metrics, our empirical validation focuses on a single dataset and model—the ISC model trained on the Leuven Concepts Database. 
This choice offers strong psychological relevance and interpretability in a controlled setting, but it leaves open important questions about how the framework scales to more complex architectures and diverse datasets. 
Furthermore, while we introduce metrics to quantify representational phenomena (e.g., manifold complexity and statistical dependence), these abstract constructs are challenging to validate and will require additional theoretical and empirical work to link them to observable behaviors in neural networks and cognitive systems. 
Finally, the complexity of approximating the AMD using a GFlowNet grows rapidly with the number of ablatable units, as longer trajectories make credit assignment and optimization more challenging, a common issue in reinforcement learning.

In conclusion, this work introduces a probabilistic framework for understanding task representations in neural networks, providing a principled approach to uncover causal relationships and representational complexity. While further development and scaling are needed, our approach lays a foundation for future research into task representations across both natural and artificial systems. We hope this framework inspires new insights into the principles governing learning and cognition in neural network-based architectures.

\acks{We thank Tyler Giallanza for providing materials related to the ISC model, Legasse Remon for assistance with running experiments, Arjun Menon for helping organize the item property taxonomy table, and Leigh Nystrom for administrative and research support.
Jonathan Cohen was supported by a Vannevar Bush Faculty Fellowship from the Office of the Under Secretary of Defense for Research \& Engineering.
This project was made possible through the support of a grant from Templeton World Charity Foundation, Inc (funder DOI 501100011730) through grant DOI.ORG/10.54224/34207. The opinions expressed in this publication are those of the author(s) and do not necessarily reflect the views of Templeton World Charity Foundation, Inc.}

\clearpage
\bibliography{main}

\clearpage
\appendix

\section{ISC Model}
\subsection{Model architecture}
The ISC model receives two one-hot vectors as inputs: an item input and a task input, with 350 and 36 possible choices, respectively. 
These one-hot vectors are passed through separate embedding layers, generating latent representations of 64 dimensions for the item input and 24 dimensions for the task input, both using sigmoid activations. 
The embeddings are then concatenated into an 88-dimensional vector and passed through a linear layer with sigmoid activations, reducing the dimensionality to produce context-dependent representations with 64 dimensions. 
The model leverages both context-independent representations (item embeddings), which represent items independently of tasks, and context-dependent representations, which incorporate task-specific information. Both context-independent (item embeddings) and context-dependent representations are mapped to the output layer, which predicts 2896 feature labels using sigmoid activations.

\subsection{Training}
The model is trained by minimizing the sum of three loss components. 
First, it computes the negative log-likelihood (NLL) for all features across tasks by taking the union of all 36 feature classes for a given item. 
Second, it computes the NLL for specific item-task combinations. 
Third, it computes the NLL for the null task, where the target outputs are all zeros. 
These three loss quantities are summed to create a balanced learning objective that captures both task-dependent and task-independent relationships.
The model is trained using the Adam \citep{kingma2014adam} optimizer with a learning rate of 0.05 over 150 epochs, corresponding to 29,550 gradient updates with a batch size of 64 item-task pairs.

\section{Numerical stability}

% \subsection{Odds-ratio}
To avoid \(-\infty\) and \(+\infty\) in the odds-ratio calculation, sensitivity and specificity are estimated using the expected value of a beta distribution. This approach ensures numerical stability, especially in cases of sparse or extreme counts. Specifically, sensitivity (\(P(\text{correct} \mid \text{task}, \text{mask}, \text{target} = 1)\)) and specificity (\(P(\text{correct} \mid \text{task}, \text{mask}, \text{target} = 0)\)) are calculated as:

\[
\text{sensitivity} = \mathbb{E}[\text{Beta}(a + 1, b + 1)]
\]
\[
\text{specificity} = \mathbb{E}[\text{Beta}(c + 1, d + 1)]
\]

\noindent where \(a\) and \(c\) represent the counts of correct predictions for positive and negative targets, respectively, and \(b\) and \(d\) represent the counts of incorrect predictions.

\section{Standard Bayes}
As noted in the main manuscript, we find that using the standard Bayesian formulation using the task accuracy directly to compute the posterior yields poorer results.
Using Bayes' theorem and assuming a uniform prior over $P(t, m)$, we have:
\[
P(m \mid t, c) = \frac{P(c \mid t, m)}{\sum_{m'} P(c \mid t, m')}
\]
We substantiate this claim by comparing how well our metrics aligns with other measures.

\subsubsection{Task differentiation}
\begin{figure}[h]
  \centering
  \includegraphics[width=\textwidth]{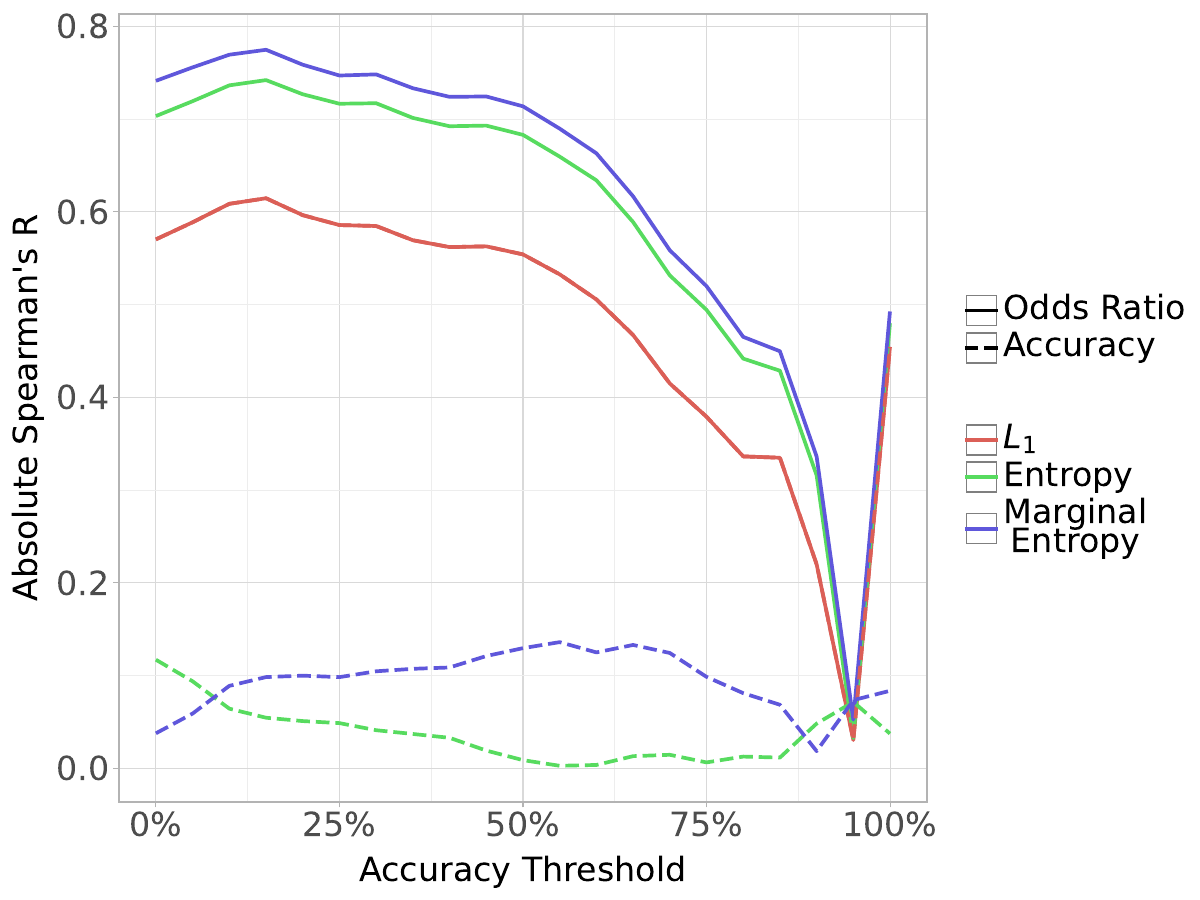}
  \caption{Spearman correlation between similarity measures.}
  \label{fig:task_rank_corr_supp}
\end{figure}

First, we examine the relationship between AMD entropy and task differentiation, as discussed in the main manuscript. Unlike the odds-ratio formulation, the standard Bayesian approach based on accuracy fails to capture the sequence in which the model begins learning each task, as shown in Figure~\ref{fig:task_rank_corr_supp}. In contrast, the odds-ratio offers a much clearer signal that closely aligns with the model’s actual learning dynamics.

\subsubsection{Task similarity}
Next, we consider the task similarity RSA, where we compare the mask-performance correlation (MPC), symmetrized KL-divergence, and the Wasserstein distance to cosine similarity and Euclidean distances.
Given the expected and observed low manifold complexity of the ISC model, cosine similarity provides a reasonable point of comparison.
Using the odds-ratio formulation, we found that the Wasserstein distance has a Spearman's correlation of 0.89 with cosine similarity, suggesting that the combination of the AMD using odds-ratio and a shape-sensitive measure like Wasserstein is able to capture similar representational structure to cosine similarity.

\begin{figure}[h]
  \centering
  \includegraphics[width=\textwidth]{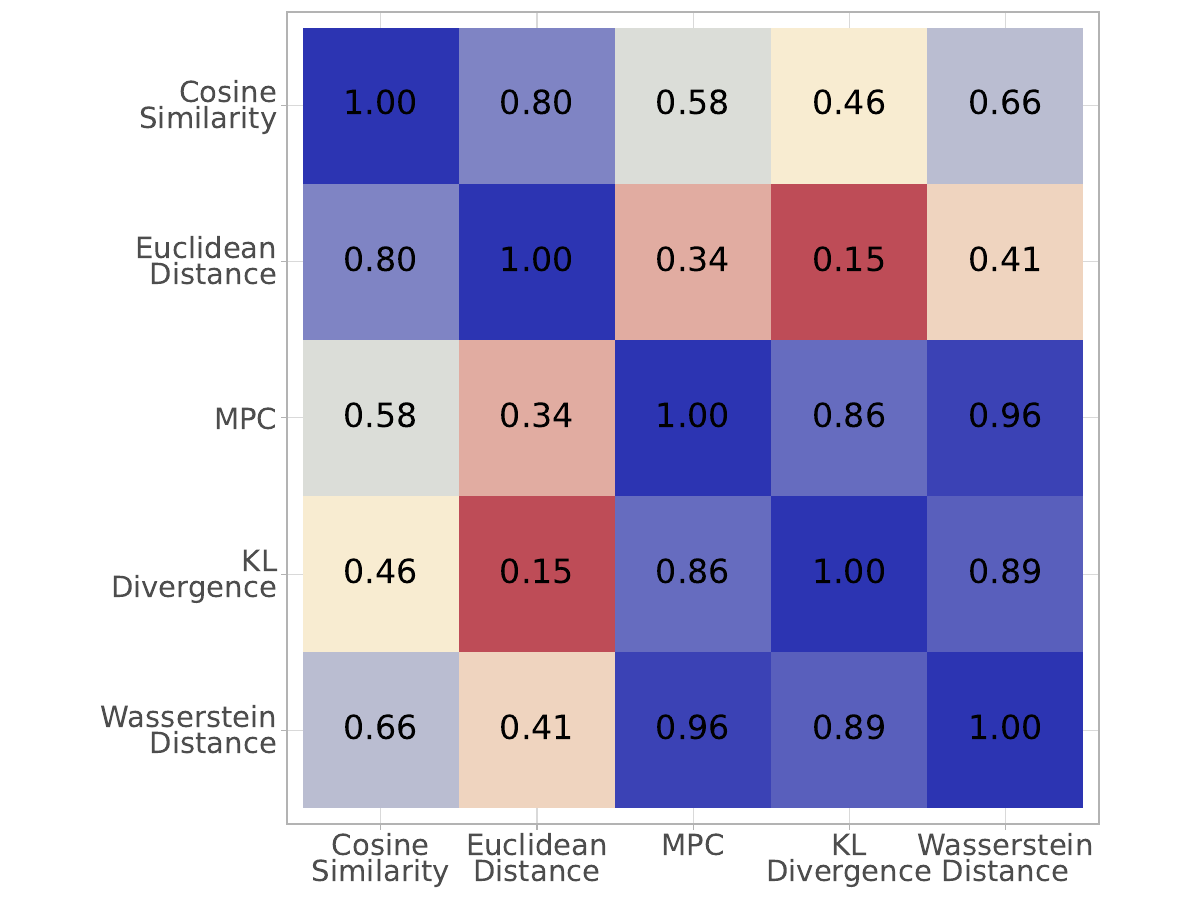}
  \caption{Spearman correlation between similarity measures.}
  \label{fig:rsa_comparison_acc}
\end{figure}
Swapping out the odds-ratio with a standard accuracy measure $P(c \mid t, m)$ to compute the AMD yields much lower correlation between the distributional measures (MPC, KL-divergence, and Wasserstein distance) and the vector-based measures (cosine similarity and Euclidean distance).
As shown in Figure~\ref{fig:rsa_comparison_acc}, while the three distributional measures are highly correlated with each other and the vector-based measures are also highly correlated with each other, the correlation between distributional measures and vector-based measures is substantially weaker.
Even Wasserstein distance, which had a 0.89 correlation with cosine similarity using the odds-ratio formulation, only has a 0.66 correlation using just the accuracy.
While 0.66 is not necessarily low correlation, it does reflect a substantial drop in alignment.

\subsection{Binomial formuation}
One possible alternative to the standard Bayesian formulation and the likelihood ratio formulation we considered is modeling task performance as a binomial distribution with $n$ independent trials, where each trial corresponds to an individual input-output pair. 
However, this formulation breaks down for large $n$ (1,013,600 in our model) where the likelihood becomes extremely peaked, causing a few masks that achieve near-perfect performance to receive equal probabilities ($1/k$, where $k$ is the number of such masks) while all other masks are driven to 0.
This sharpness effectively collapses the range of possible outcomes and reduces the ability to distinguish between masks with subtle differences in performance.

\section{GFlowNet}

Our GFlowNet model is implemented as a multilayer perceptron (MLP) with three hidden layers, each containing 1024 units and using the Exponential Linear Unit (ELU) \citep{clevert2016elu} activation function. The model takes the current mask state \( m^{(j)} \) as input and outputs a vector of size \( 2d + 1 \), which is decoded into the forward policy \( P_\theta^f (m^{(j+1)} \mid m^{(j)}; t) \), including the termination probability \( P_\theta(\top \mid m^{(j)}, t) \), and the backward policy \( P_\theta^b (m^{(j)} \mid m^{(j+1)}; t) \).  

To facilitate training, we maintain a replay buffer that stores the first \( 0.01 \cdot 2^{24} = 167,772 \) masks explored by the GFlowNet. Every 100 gradient updates, we sample 250 trajectories using off-policy exploration, where the next state is selected uniformly at random with a 5\% probability. For each gradient update, 1000 transitions are drawn from the replay buffer, and the model parameters are optimized using the Adam optimizer \citep{kingma2014adam} with a learning rate of 0.001.

\section{Tables and figures}
\begin{figure*}[h]
  \centering
  \includegraphics[width=1\textwidth]{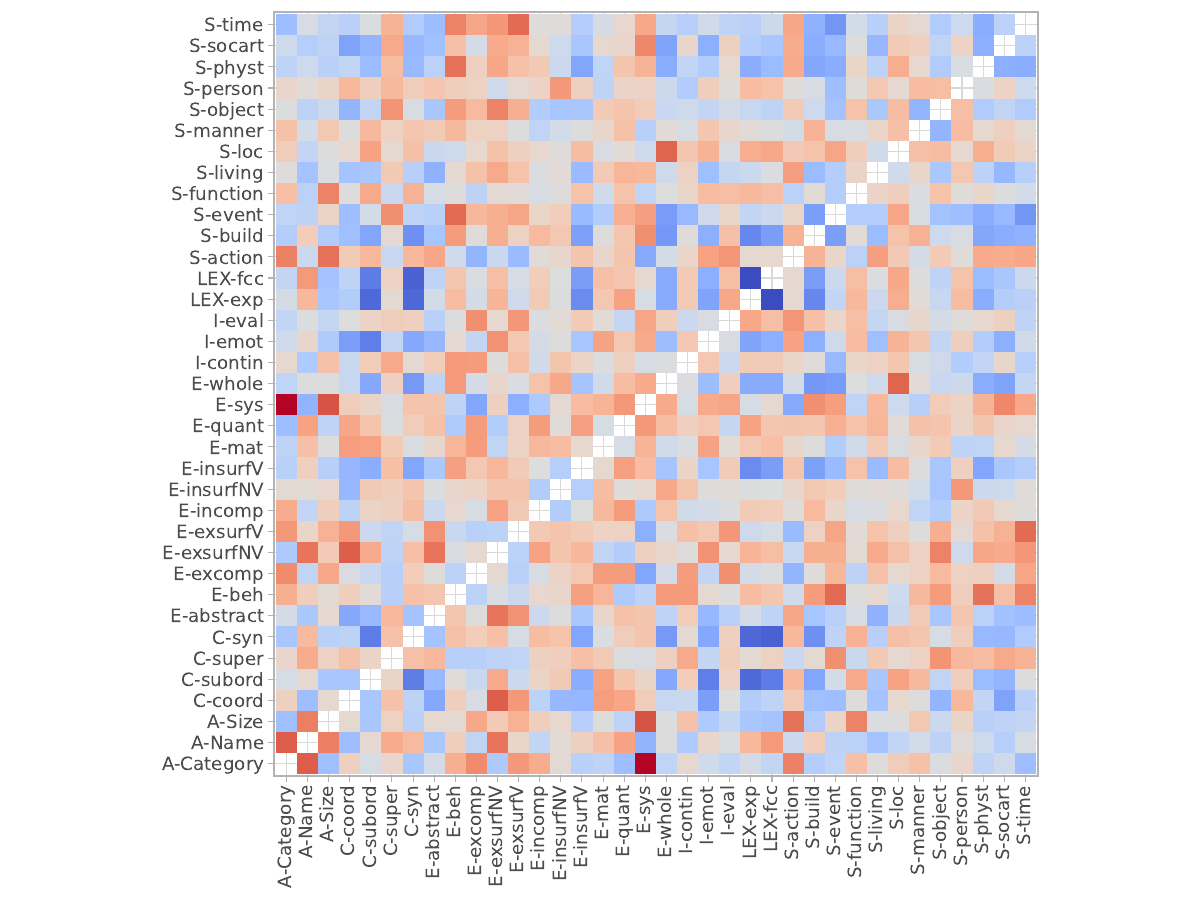}
  \caption{Cosine distance between tasks.}
  \label{fig:cosine_supp}
\end{figure*}
\clearpage

\begin{figure*}[h]
  \centering
  \includegraphics[width=1\textwidth]{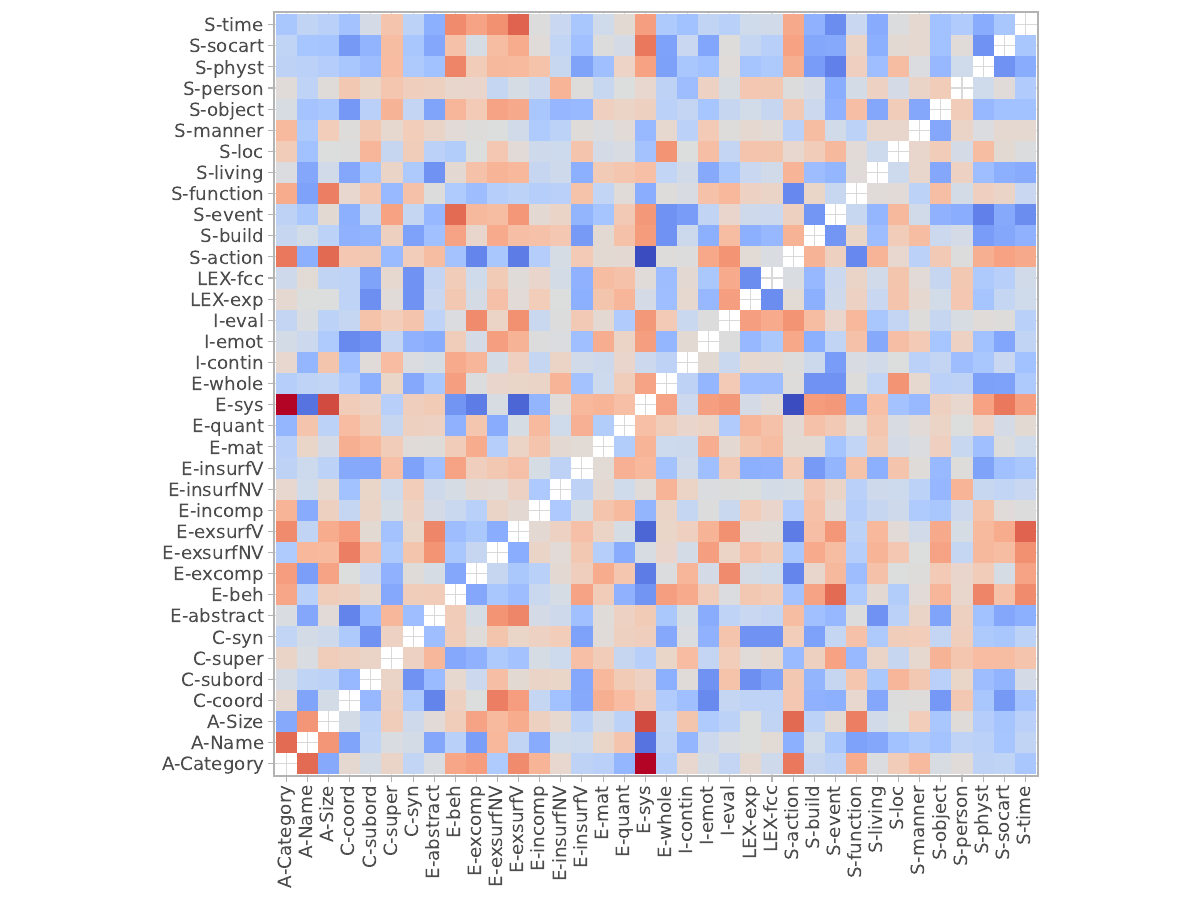}
  \caption{Wasserstein distance between tasks.}
  \label{fig:wasserstein_supp}
\end{figure*}
\clearpage

\begin{landscape}

\begin{table}[ht]
\centering
\caption{Taxonomy of item properties used to generate tasks for the model. The ``Additional Tasks'' types were added for the simulations in \cite{giallanza2024integrated} and the remaining 33 property types were coded according to Wu and Barsalou’s taxonomy \cite{wu2009perceptual}. Adapted from \cite{giallanza2024integrated}.}
\label{table:dataset_property_types}

% ----------- TABULAR 1: Additional + Taxonomic ----------- 
\vspace{0.5em}
\begin{tabular}{|p{2.5cm}|p{3cm}|p{2cm}|p{7cm}|p{5cm}|}
\hline
\textbf{Superordinate Property Type} & \textbf{Subordinate Property Type} & \textbf{Task ID} & \textbf{Description} & \textbf{Example} \\
\hline

% Taxonomic Category (4 rows)
\multirow{4}{*}{\begin{tabular}[c]{@{}l@{}}Taxonomic\\Category\end{tabular}} 
    & Coordinate Category    & C-coord  & A different concept belonging to the same category as the target concept & ``Dogs'' are like coyotes \\ \cline{2-5}
    & Subordinate Category   & C-subord & A category one level below the target concept's category & ``Dog'' can be a beagle \\ \cline{2-5}
    & Superordinate Category & C-super  & A category one level above the target concept’s category & ``Dogs'' are mammals \\ \cline{2-5}
    & Synonym                & C-syn    & A synonym of the concept & ``Dog'' is the same as canine \\ \hline

% Lexical Attributes (2 rows)
\multirow{2}{*}{\begin{tabular}[c]{@{}l@{}}Lexical\\Attributes\end{tabular}} 
    & Expression        & LEX-exp & Words that occur in commonly used expressions & ``Birds and the bees''; ``Bird in the hand is worth two in the bush'' \\ \cline{2-5}
    & Forward Completion & LEX-fcc & The use of a word as a prefix to the response & ``Bomb'' shelter; ``Monkey'' business \\ \hline

% Introspective Attributes (3 rows)
\multirow{3}{*}{\begin{tabular}[c]{@{}l@{}}Introspective\\Attributes\end{tabular}} 
    & Contingency    & I-contin & A contingency between the concept and a situational aspect, such as causation, correlations, dependency, etc. & ``Garlic'' causes bad breath \\ \cline{2-5}
    & Evaluation     & I-eval   & A positive or negative evaluation of the concept or its components & ``Tuxedos'' are fancy; ``Apples'' taste good \\ \cline{2-5}
    & Affect/Emotion & I-emot   & An emotional state associated with perceiving the concept & ``Wasps'' are annoying \\ \hline

% Additional Tasks (3 rows)
\multirow{3}{*}{\begin{tabular}[c]{@{}l@{}}Additional\\Tasks\end{tabular}} 
    & Animal vs Instrument & A-Category & The basic-level category of the object & ``Elephants'' are animals \\ \cline{2-5}
    & Name                 & A-Name     & The name of the concept & ``Dogs'' are called ``dog'' \\ \cline{2-5}
    & Size                 & A-Size     & Whether the object is larger than a folding chair & ``Elephants'' are large \\ \hline

\end{tabular}
\end{table}

% ----------- TABULAR 2: Entity ----------- 
\begin{table}[ht]
\centering
\vspace{0.5em}
\begin{tabular}{|p{2.5cm}|p{3cm}|p{2cm}|p{7cm}|p{5cm}|}
\hline
\textbf{Superordinate Property Type} & \textbf{Subordinate Property Type} & \textbf{Task ID} & \textbf{Description} & \textbf{Example} \\
\hline

% Entity Attributes (12 rows)
\multirow{12}{*}{\begin{tabular}[c]{@{}l@{}}Entity\\Attributes\end{tabular}} 
    & Abstract                & E-abstract & An abstract property of the concept & ``Simon'' is a democrat \\ \cline{2-5}
    & Behavior                & E-beh      & A behavior characteristic of the concept & ``Dogs'' bark \\ \cline{2-5}
    & External Component      & E-excomp   & A component of the concept that resides on its exterior or surface & ``Dogs'' wear collars \\ \cline{2-5}
    & External Surface Feature (Non-Visual) & E-exsurfNV & An external feature of a concept that is not a component and is not visual (e.g., touch, smell, taste) & ``Dogs'' feel fuzzy \\ \cline{2-5}
    & External Surface Feature (Visual)     & E-exsurfV  & An external feature of a concept that is not a component and is visual (e.g., shape, color, texture) & ``Dogs'' are brown or white \\ \cline{2-5}
    & Internal Component      & E-incomp   & A component of the concept that resides within the interior of the concept & ``Dogs'' have hearts \\ \cline{2-5}
    & Internal Surface Feature (Non-Visual) & E-insurfNV & An internal feature of the concept that is not a component, not visible, and perceived when the concept’s interior is exposed & ``Apples'' taste sweet \\ \cline{2-5}
    & Internal Surface Feature (Visual)     & E-insurfV  & An internal feature of the concept that is not a component, is visible, and is perceived when the concept’s interior is exposed & ``Watermelons'' are red on the inside \\ \cline{2-5}
    & Material                & E-mat      & The material the object is made out of & ``Dressers'' are made from wood \\ \cline{2-5}
    & Quantity                & E-quant    & A numerosity, frequency, or intensity characteristic of the concept & ``Giraffes'' have long necks \\ \cline{2-5}
    & Systemic                & E-sys      & A systemic feature of the concept or its components, including states, conditions, abilities, and traits & ``Dolphins'' are intelligent \\ \cline{2-5}
    & Larger Whole            & E-whole    & A whole to which an entity belongs & ``Ants'' are part of colonies \\ \hline

\end{tabular}
\end{table}

% ----------- TABULAR 3: Situational ----------- 
\begin{table}[ht]
\centering
\vspace{0.5em}
\begin{tabular}{|p{2.5cm}|p{3cm}|p{2cm}|p{7cm}|p{5cm}|}
\hline
\textbf{Superordinate Property Type} & \textbf{Subordinate Property Type} & \textbf{Task ID} & \textbf{Description} & \textbf{Example} \\
\hline

% Situational Attributes (12 rows)
\multirow{12}{*}{\begin{tabular}[c]{@{}l@{}}Situational\\Attributes\end{tabular}} 
    & Action         & S-action  & An action that an agent performs with or relative to the target concept & ``Apples'' are picked; ``Shirts'' are worn \\ \cline{2-5}
    & Building       & S-build   & A building associated with the target concept & ``Books'' are found in libraries; ``Pipe Organs'' are found in churches \\ \cline{2-5}
    & Event          & S-event   & An event commonly associated with the target object & ``Roses'' are given as gifts on Valentine’s Day \\ \cline{2-5}
    & Function       & S-function & A goal the concept is used to achieve or a function the concept is used for & ``Apples'' are eaten \\ \cline{2-5}
    & Living         & S-living  & A living thing in a situation that is not a person, including plants and other animals & ``Beds'' are used by dogs; ``Apples'' grow on trees \\ \cline{2-5}
    & Location       & S-loc     & A location where the target concept can be found & ``Zebras'' are found in Africa \\ \cline{2-5}
    & Manner         & S-manner  & The manner in which an action involving the concept is performed & ``Potatoes'' are cooked; ``Eggs'' are turned into omelettes \\ \cline{2-5}
    & Object         & S-object  & A concrete object commonly associated with the target concept & ``Carrots'' are cooked in a pot \\ \cline{2-5}
    & Person         & S-person  & A person or group of people in a situation & ``Toys'' are used by children; ``Cars'' contain passengers and drivers \\ \cline{2-5}
    & Physical State & S-physt   & A physical state of the concept or a component of the concept & ``Mountains'' are snowy \\ \cline{2-5}
    & Social Artifact & S-socart & A relatively abstract entity created by socio-cultural institutions that relates to the concept (e.g., a book or a movie) & ``Whales'' are the subject of Moby Dick \\ \cline{2-5}
    & Time           & S-time    & A period of time associated with the concept & ``Turkeys'' are eaten in November; ``Apples'' are in season in Autumn \\ \hline
\end{tabular}

\end{table}
\end{landscape}

\end{document}